\documentclass[lettersize,journal]{IEEEtran}
\usepackage{amsmath,amsfonts}
\usepackage{algorithmic}
\usepackage{algorithm}
\usepackage{array}
\usepackage[caption=false,font=normalsize,labelfont=sf,textfont=sf]{subfig}
\usepackage{textcomp}
\usepackage{stfloats}
\usepackage{url}
\usepackage{verbatim}
\usepackage{graphicx}
\usepackage{cite}
\usepackage{pifont}
\usepackage{bbm}
\usepackage{multirow}

\begin{document}

\title{PM-VIS: High-Performance Box-Supervised Video Instance Segmentation}

\author{Zhangjing Yang, Dun Liu, Wensheng Cheng, Jinqiao Wang, Yi Wu$^{*}$

\thanks{
Zhangjing Yang and Dun Liu are with the School of College of Computer Science, Nanjing Audit University,  Nangjing, China. E-mail: yzj@nau.edu.cn, einstonliu@163.com.

Wensheng Cheng is with the Department of Computer Science, Stony Brook University, NY 11794, United States. E-mail: wenscheng@cs.stonybrook.edu.

Jinqiao Wang is with the Institute of Automation, Chinese Academy of Sciences, 95 Zhongguancun East Road, Beijing, 100190, China. E-mail: jqwang@nlpr.ia.ac.cn.

Yi Wu is with Lucid Motors, 7373 Gateway Blvd, Newark, CA 94560, United States. E-mail: ywu.china@gmail.com. $^{*}$Corresponding Author.
}
}



\maketitle

\begin{abstract}
Labeling pixel-wise object masks in videos is a resource-intensive and laborious process. 
Box-supervised Video Instance Segmentation (VIS) methods have emerged as a viable solution to mitigate the labor-intensive annotation process. 
However, previous approaches have primarily emphasized a traditional single-step strategy, overlooking the potential of a two-step strategy involving the use of segmentation models to generate pseudo masks, and subsequently leveraging them alongside box annotations for model training. In practical applications, the two-step approach is not only more flexible but also exhibits a higher recognition accuracy.
Inspired by the recent success of Segment Anything Model (SAM), we introduce a novel approach that aims at harnessing instance box annotations from multiple perspectives to generate high-quality instance pseudo masks, thus enriching the information contained in instance annotations. 
We leverage ground-truth boxes to create three types of pseudo masks using the HQ-SAM model, the box-supervised VIS model (IDOL-BoxInst), and the VOS model (DeAOT) separately, along with three corresponding optimization mechanisms. 
Additionally, we introduce two ground-truth data filtering methods, assisted by high-quality pseudo masks, to further enhance the training dataset quality and improve the performance of fully supervised VIS methods.
To fully capitalize on the obtained high-quality \textbf{P}seudo \textbf{M}asks, we introduce a novel algorithm, PM-VIS, to integrate mask losses into IDOL-BoxInst. Our PM-VIS model, trained with high-quality pseudo mask annotations, demonstrates strong ability in instance mask prediction, achieving state-of-the-art performance on the YouTube-VIS 2019, YouTube-VIS 2021, and OVIS validation sets, notably narrowing the gap between box-supervised and fully supervised VIS methods. Furthermore, the PM-VIS model trained on these filtered ground-truth data shows obvious improvement compared to the baseline algorithm.
\end{abstract}



\section{Introduction}
\IEEEPARstart{V}{ideo} Instance Segmentation (VIS) aims to simultaneously detect, segment, and track objects within a video. Although current fully supervised VIS algorithms~\cite{Mask2formerVIS,IDOL,Vita,GenVIS,GRAtt-VIS} have demonstrated impressive performance on various benchmarks~\cite{VIS,YTVIS2021,OVIS}, their practical applications still face significant challenges. A fundamental reason is the substantial cost and difficulty on obtaining pixel-level annotations, in contrast to the less demanding bounding box or point annotations. Moreover, existing datasets often fail to capture the diverse scenarios encountered in real-world applications. Therefore, our approach focuses on training model with bounding box annotations only, enabling low-cost pixel-level instance predictions.
\begin{table}[t]
    \small
	\centering
	\renewcommand\arraystretch{1.3}
	\setlength\tabcolsep{4pt} 
	\caption{Comparison of mask AP across the three box-supervised methods MaskFreeVIS, IDOL-BoxInst, and PM-VIS on YTVIS2019, YTVIS2021, and OVIS. Results are reported with ResNet-50 as the backbone. ``Sup." indicates algorithm supervision condition.}
	\label{table:introVS}
\begin{tabular}{l|l|l|l|l}
\hline
Method        & Sup.  & YTVIS2019  & YTVIS2021  & OVIS     \\ \hline\hline
MaskFreeVIS~\cite{maskfreeVIS} & Box   & 46.6     & 40.9 & 15.7  \\
IDOL-BoxInst   & Box   & 43.9 & 41.8 & 25.4 \\
PM-VIS & Box   & \textbf{48.7} & \textbf{44.6} &\textbf{27.8}\\ \hline
\end{tabular}
\end{table}

Building upon the success of box-supervised Image Instance Segmentation (IIS) algorithms~\cite{Boxteacher,BoxCaseg,BoxInst,DiscoBox}, the sole box-supervised VIS method~\cite{maskfreeVIS} has shown promising results. 
However, prior methods have predominantly favored a traditional single-step strategy, directly employing ground-truth boxes (gtboxes) for model training. This approach overlooks the potential of a two-step strategy, which incorporates segmentation models to generate pseudo masks, subsequently leveraging them in conjunction with box annotations during model training.
Expanding upon the single-step method, the two-step approach incorporates more instance information from multiple perspectives, strengthening the algorithm's recognition capabilities.
In practical applications, the two-step approach not only enhances flexibility but also demonstrates superior recognition accuracy, significantly reducing the application costs of VIS and accelerating their development.

\begin{figure*}[!h]
	\centering
	\scalebox{0.2}{\includegraphics{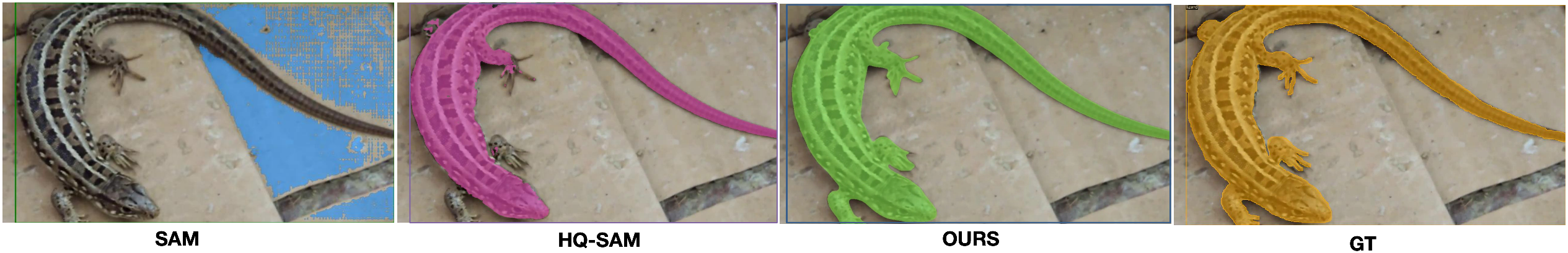}}
	\caption{Comparing pixel-level quality visualization results among the pseudo mask (SAM) from SAM-Masks, the pseudo mask (HQ-SAM) from HQ-SAM-masks, the pseudo mask (OURS) from Track-masks-final, and the mask (GT) from gtmasks.}
	\label{fig:SAMVSHQSAM-vis} 
\end{figure*}

\begin{figure}[h]
	\centering
	\scalebox{0.10}{\includegraphics{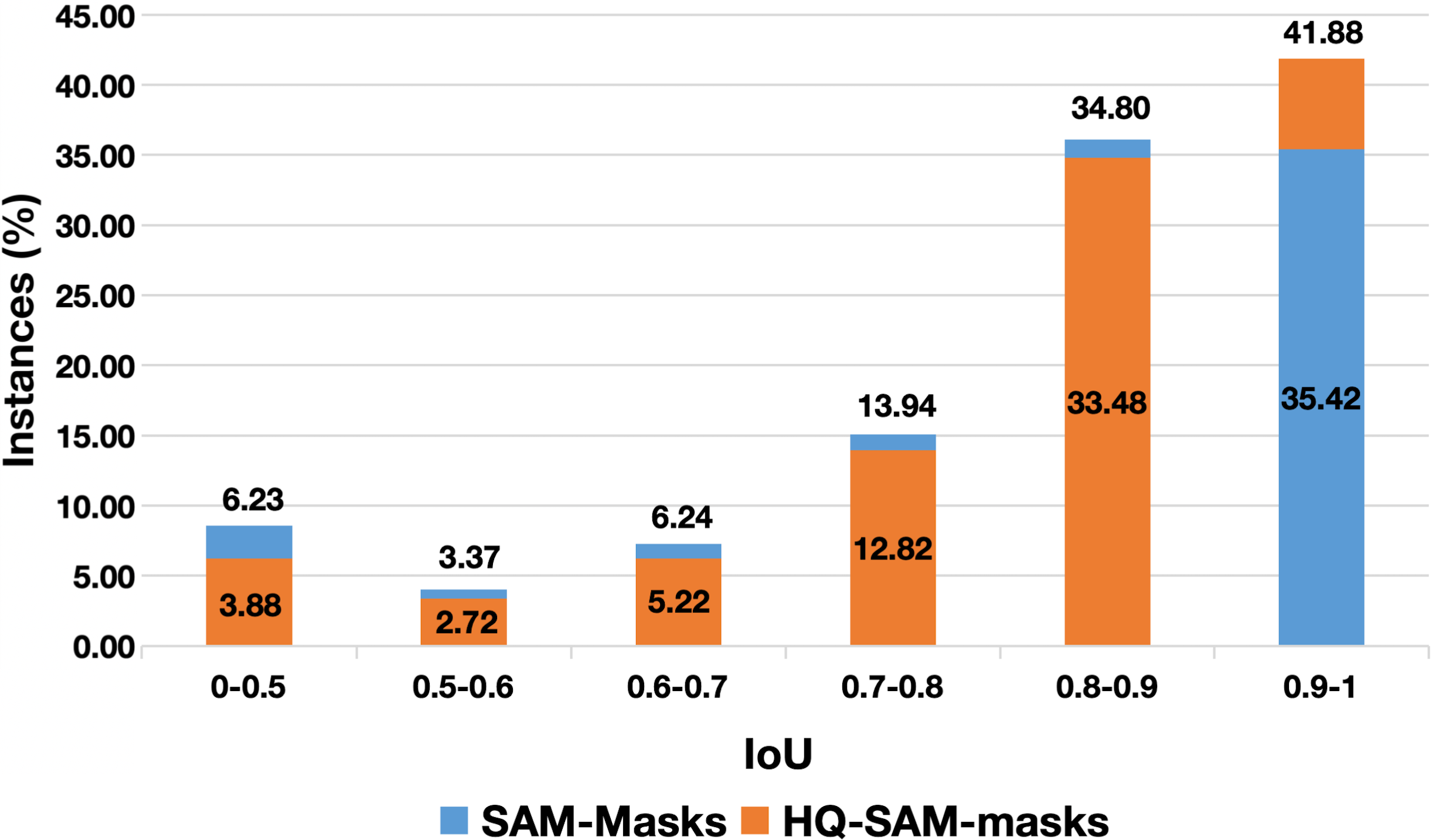}}
	\caption{The distribution of IoU between SAM-Masks and gtmasks, as well as between HQ-SAM-masks and gtmasks. The horizontal and vertical axes represent the IoU ranges and the percentage of instances within the range, respectively.}
	\label{fig:SAMVSHQSAM-fig} 
\end{figure}
\begin{figure}[!t]
	\centering
	\scalebox{0.39}{\includegraphics{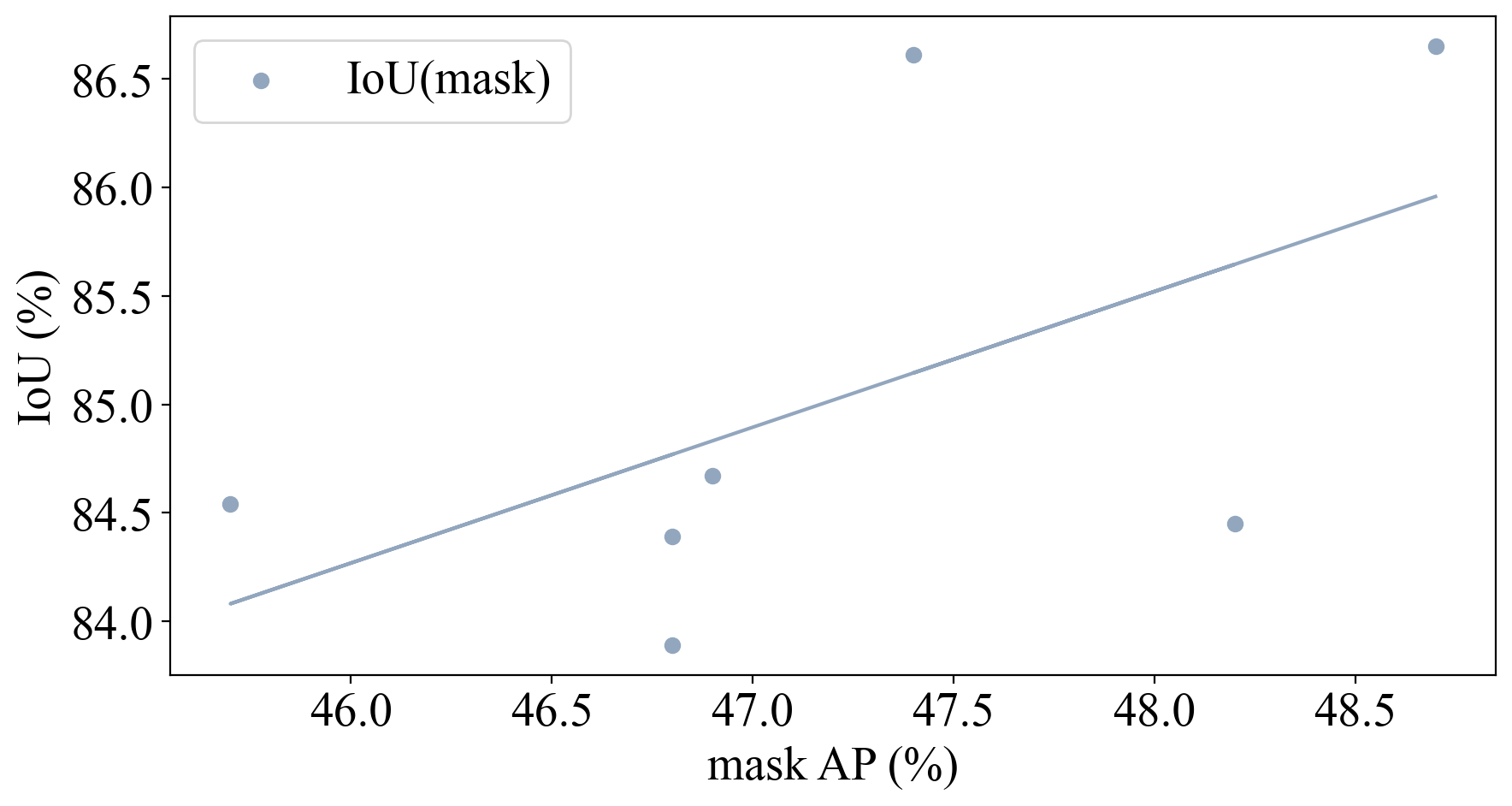}}
	\caption{The correlation between mask IoU and algorithm mask AP for pseudo masks and gtmasks. The PM-VIS model is trained on HQ-SAM-masks, IDOL-BoxInst-masks, and Track-masks, along with their derived pseudo mask collections, and its performance is evaluated in terms of mask AP on the YTVIS2019 validation set.}
	\label{fig:correlationIOUwithAP} 
\end{figure}

Taking inspiration from the success of foundational models such as SAM~\cite{Segmentanything} and HQ-SAM~\cite{HQSAM} in image segmentation, we propose to utilize high-performance models to generate target-specific pixel-level predictions (pseudo masks) based on object box proposals.
Then these pseudo masks are introduced as additional mask annotations of the training set.
However, this direct utilization is problematic.
Through visualization in Fig.~\ref{fig:SAMVSHQSAM-vis}, we have identified certain outrageous errors within the generated pseudo-masks.
While the SAM model excels in image segmentation, it faces challenges in scenarios where objects share colors with the background, have distinct background features, or exhibit complex color compositions within objects, as shown in Fig.~\ref{fig:SAMVSHQSAM-vis} (SAM).
Quantitatively, through intersection over union (IoU) calculations between the pseudo masks (SAM-Masks) generated by SAM and the ground-truth masks (gtmasks), as shown in Fig.~\ref{fig:SAMVSHQSAM-fig}, we observe that the SAM-Masks contain over 64.58\% lower-quality data, with an IoU score less than 0.9. Therefore, we believe directly using these noisy data will adversely affect the performance of weakly supervised methods.



To obtain superior-quality pseudo masks and fully harness their potential, our method involves three key components: \textbf{(i)}~generating three types of pseudo masks using distinct models, \textbf{(ii)}~selecting high-quality pseudo masks from the aforementioned three using SCM, DOOB, and SHQM strategies, and \textbf{(iii)}~training the proposed PM-VIS model on the acquired high-quality pseudo masks. 
While the pseudo masks data has enhanced the performance of weakly supervised algorithms, we believe that there is still untapped useful information within them. Therefore, we introduce an effective approach to fully explore and leverage the hidden information in pseudo mask data: 
\textbf{(i)}~filtering ground-truth data using the obtained high-quality pseudo masks, and \textbf{(ii)}~training the proposed PM-VIS model on the filtered ground-truth data.

\subsection{Pseudo Mask Generation} 
To effectively leverage information from gtboxes, we propose utilizing three models (HQ-SAM, IDOL-BoxInst, and DeAOT) to generate instance pseudo masks: HQ-SAM-masks, IDOL-BoxInst-masks, and Track-masks.
\subsubsection{HQ-SAM-masks} 
In order to mitigate the noise in the generated pseudo masks, we employ HQ-SAM, which incorporates a learnable high-quality output token, to produce these pseudo masks, referred to as HQ-SAM-masks. The HQ-SAM model, as shown in Fig.~\ref{fig:SAMVSHQSAM-vis} (HQ-SAM), significantly reduces the number of outrageous errors compared to SAM.
Besides, as depicted in Fig.~\ref{fig:SAMVSHQSAM-fig}, by calculating the IoU between HQ-SAM-masks and gtmasks, we find that HQ-SAM-masks exhibits 6.46\% higher high-quality (IoU greater than 0.9) instance count compared to SAM-Masks and has a lower instance count across all intervals of IoU less than 0.9 compared to SAM-Masks.

\subsubsection{IDOL-BoxInst-masks}
We observe that HQ-SAM-masks still exhibits diminished quality in scenarios involving occlusion and rapid motion.
Taking into account the effectiveness of the IDOL algorithm in handling occlusion, we introduce IDOL-BoxInst-masks, a set of pseudo masks generated by the box-supervised VIS model IDOL-BoxInst. It combines box-supervised IIS model BoxInst~\cite{BoxInst} with VIS model IDOL~\cite{IDOL} while excluding mask losses~\cite{DiceLoss,FocalLoss} from IDOL. This model is trained on the dataset without mask annotations.

\subsubsection{Track-masks}
Despite aforementioned techniques, pseudo masks generated by weakly supervised model still struggle to guarantee the accuracy and completeness of object predictions, especially in scenes with subtle color variations. 
We observe that within HQ-SAM-masks and IDOL-BoxInst-masks, despite the presence of some subpar segmentations, there are many high-quality instance segmentations that reach the ground-truth level.
Subsequently, we employ the semi-supervised Video Object Segmentation (VOS) model DeAOT~\cite{DeAOT} with initialization of object annotations that comes from the high-quality pseudo masks in IDOL-BoxInst-masks, to track the pseudo masks of the target throughout the entire video, resulting in a novel pseudo mask collection called Track-masks. 

\subsection{Pseudo Mask Selection}
To fully leverage the advantages of various pseudo masks, we propose three strategies: SCM, DOOB, and SHQM. These strategies are designed to optimize the quality of pseudo masks and select those of higher quality.
\subsubsection{SCM}
During the inference stage of the IDOL-BoxInst, the model predicts multiple masks for each real instances. However, it is challenging to ascertain the quality of each predicted mask. 
Using predicted target labels or IDs alone does not uniquely establish the relationship with instance gtboxes. 
To establish correspondences between predicted masks and gtboxes, we propose a method that utilizes both the IoU between gtboxes and predicted boxes, as well as the IoU between HQ-SAM-masks and predicted masks to determine their associations.
Specifically, we utilize HQ-SAM-masks for temporal continuity and object matching, constraining the predicted pseudo masks using our proposed \textbf{SCM} (\textbf{S}coring \textbf{C}orresponding \textbf{M}ask) confidence scores.

\subsubsection{DOOB}
Upon closer examination of these segmented instances, two distinct issues within the segmented instance annotations appear: \textbf{(i)} overlapping occurs between distinct predicted masks on the same frame, and \textbf{(ii)} the predicted masks extend beyond the boundaries of the corresponding gtbox. To address these concerns, we introduce a pseudo mask optimization strategy, \textbf{DOOB} (\textbf{D}eleting the \textbf{O}verlapping and \textbf{O}ut-of-\textbf{B}oundary sections). Specifically, DOOB involves the removal of overlapping portions of pseudo masks from different instances and the removal of parts extending beyond the gtbox boundaries.

\subsubsection{SHQM}
With our multi-keyframes tracking strategy, we obtain a combination of high-quality pseudo masks, named Track-masks-final, selected from the three aforementioned sets (HQ-SAM-masks, IDOL-BoxInst-masks, Track-masks) using \textbf{SHQM} (\textbf{S}election of \textbf{H}igh-\textbf{Q}uality \textbf{M}asks). Specifically, the SHQM method determines the higher-quality pseudo mask among the three by summing the SCM scores calculated between the target pseudo-mask and the other two pseudo-masks, where a higher SCM score indicates higher quality. As illustrated in Fig.~\ref{fig:SAMVSHQSAM-vis}~(OURS), Track-masks-final exhibits significantly improved visual quality compared to the previous pseudo masks, closely resembling the gtmasks.

\subsection{Ground-Truth Data Filtration}

We introduce two ground-truth data filtering methods: \textbf{Missing-Data} (Removing \textbf{Missing Data} from ground-truth) and \textbf{RIA} (\textbf{R}emoving \textbf{I}nstance \textbf{A}nnotations with low mask IoU between Track-masks-final and gtmasks).
Specifically, we enhance our results through two perspectives. Initially, using the Missing-Data method, we improve the result quality by removing missing data mappings from Track-masks-final in the ground-truth data. 
We then utilize the problematic pseudo masks from Track-masks-final to further optimize the ground-truth data using the RIA method.

\subsection{PM-VIS}
To fully explore the information from both boxes and \textbf{P}seudo \textbf{M}asks, we combine IDOL-BoxInst with mask losses from IDOL~\cite{IDOL}, resulting in the novel \textbf{VIS} algorithm, \textbf{PM-VIS}.
Generally, for PM-VIS, the higher the quality of pseudo masks, i.e., the closer they are to manually annotated masks, the better the algorithm performance is expected to be. 
As shown in Fig.~\ref{fig:correlationIOUwithAP}, we compute the mask IoU between each type of pseudo masks and the gtmasks, and plot the correlation between mask IoU and algorithm mask AP performance. 
There is a direct correlation between the mask IoU of pseudo masks and the performance of mask AP.
When the mask IoU between datasets containing pseudo masks and ground-truth data increases, the quality of pseudo masks improves, leading to better performance for PM-VIS on datasets with a higher mask IoU.
Therefore, training the PM-VIS model with the highest-quality pseudo masks obtained from Track-masks-final will result in the best AP performance.



As indicated in Table~\ref{table:introVS},  training the PM-VIS model on the training set containing Track-masks-final, without using gtmasks, results in state-of-the-art (SOTA) AP scores of 48.7\%, 44.6\%, and 27.8\% on the YouTube-VIS 2019~\cite{VIS}, YouTube-VIS 2021~\cite{YTVIS2021}, and OVIS~\cite{OVIS} validation sets, respectively, with the ResNet-50 backbone.  It's worth noting that the results for MaskFreeVIS~\cite{maskfreeVIS} are a combination of the Mask2Former~\cite{Mask2Former} baseline on YTVIS2019/2021 and the VITA~\cite{Vita} baseline on OVIS. 
Meanwhile, we achieve competitive performance when training the fully supervised PM-VIS model with the filtered ground-truth data, surpassing the baseline algorithm (IDOL).

The paper presents the following contributions:
\begin{itemize}
  \item We introduce three methods for generating pseudo masks. Among these, we propose, for the first time, the use of the box-supervised model (IDOL-BoxInst) to generate pseudo masks for the training set, where IDOL-BoxInst-masks plays a crucial role in producing high-quality pseudo masks.
  \item We introduce three strategies, SCM, SHQM, and DOOB, for the selection and composition of high-quality pseudo masks from the aforementioned three types of pseudo masks.
  \item We propose two ground-truth data filtering methods, Missing-Data and RIA, assisted by Track-masks-final, to enhance the quality of ground-truth data through the utilization of pseudo masks.
  \item Utilizing the box-supervised PM-VIS model with training data incorporating high-quality pseudo masks, we achieve state-of-the-art results across the YTVIS2019, YTVIS2021, and OVIS datasets. 
Meanwhile, our fully supervised PM-VIS model, trained on meticulously filtered ground-truth data across these three datasets, not only surpasses the baseline (IDOL), but also demonstrates performance on par with strong fully supervised algorithms.
\end{itemize}

\section{Related work}
\label{sec:related-work}

\subsection{Video Instance Segmentation}	
In the field of VIS, methods can be categorized into two types based on their application modes: offline and online.
Offline methods~\cite{Tube-Link,Seqformer,Vita,MaskProp,propose-reduce,TCSVT-VIS-offline} analyze the entirety of a video or a video segment, given the availability of future frames during the inference phase. 
These techniques are often suitable for tasks involving post-processing and batch analysis, including video editing, content analysis, and offline video comprehension.
Early offline methods~\cite{MaskProp,propose-reduce} use instance mask propagation to track the same target across the sequence.
With the advent of DETR~\cite{detr}, instance query tracking and matching-based VIS methods~\cite{Seqformer,Vita,Tube-Link} have gained popularity due to their remarkable effectiveness.
Online methods~\cite{GRAtt-VIS,Minvis,IDOL,VIS,TCSVT-VIS-online} process instance segmentation and tracking results for each target in video frames by embedding similarity between targets during tracking and optimizing tracking outcomes. 
These approaches find extensive applications in domains such as video surveillance and autonomous driving. 
For instance, Mask-Track R-CNN~\cite{VIS}, an extension of the IIS model Mask R-CNN~\cite{maskRCNN}, associates instances with track heads during inference. 
Before the introduction of IDOL~\cite{IDOL}, online methods typically demonstrated lower performance in comparison to offline methods. IDOL enhances conventional techniques through the incorporation of contrastive learning on instance queries across frames and the adoption of heuristic instance matching, building upon Deformable-DETR~\cite{deformable-detr}.
MinVIS~\cite{Minvis} harnesses the capabilities of the powerful object detector Mask2Former~\cite{Mask2Former} and employs bipartite matching for instance tracking. 
GenVIS~\cite{GenVIS} elevates performance through a strategic video-level training approach centered on the Mask2Former~\cite{Mask2Former}. 
In consideration of algorithm simplicity and robustness, we propose two weakly supervised VIS algorithms, namely, IDOL-BoxInst and PM-VIS, based on the foundation of IDOL.

\subsection{Video Object Segmentation}
VOS methods can be categorized into two types based on how the corresponding object is determined: automatic VOS and semi-automatic (or semi-supervised) VOS. Automatic VOS methods~\cite{zhou2020motionAutoVOS,siam2023multiscaleAutoVOS,ren2021reciprocalAutoVOS} do not require initial object annotations, while semi-supervised VOS methods~\cite{TCSVT-VOS1,cheng2022xmem,DeAOT,AOT,TCSVT-VOS3} necessitate providing annotations for the target, typically in the form of annotations in the first frame. Subsequently, the algorithm tracks the corresponding object throughout the entire video.
For semi-supervised VOS, AOT~\cite{AOT} proposes to incorporate an identification mechanism to associate multiple objects in a unified embedding space which enables it to handle multiple objects in a single propagation. 
DeAOT~\cite{DeAOT} is a semi-supervised VOS method that utilizes an AOT-inspired hierarchical propagation strategy. This strategy incorporates a dual-branch gated propagation module (GPM) to distinctly guide the propagation of visual and identification embeddings. It helps mitigate the potential loss of object-agnostic visual details in deeper layers. Furthermore, DeAOT demonstrates remarkable competence in effectively handling small or dynamically scaled objects. Given its superior performance and robustness, we employ DeAOT as our tracking model to generate pseudo mask annotations for the target throughout the video. This is done by leveraging existing mask annotations provided for the target within the video. It's important to note that during the process of obtaining instance pseudo masks through tracking, we cannot always guarantee the availability of initial frame information for each instance. Therefore, in this study, we adapt DeAOT to be more flexible, allowing us to initialize instances from various positions within the video and subsequently track them throughout the entire video.

\subsection{Box-supervised Video Instance Segmentation}	
As the sole box-supervised VIS algorithm, MaskFreeVIS~\cite{maskfreeVIS} has been inspired by box-supervised IIS and has incorporated the model~\cite{BoxInst} into the pixel-supervised VIS baseline. It maintains competitive performance by introducing temporal or spatial correlation losses. Diverging from the aforementioned box-supervised VIS methodology, our contribution, PM-VIS, adopts a more integrative approach to tackle prevailing challenges by synergizing box-supervised IIS algorithm with the VIS task. It employs a training set containing pseudo masks, effectively combines the strengths of these algorithmic paradigms, and achieves promising performance on multiple public benchmarks.

\subsection{Segment Anything models}	
SAM~\cite{Segmentanything} is a pioneering model in the field of interactive image segmentation, showcasing remarkable zero-shot generalization capabilities and the ability to generate high-quality masks from just a single foreground point.
Extending this innovation to intricate object structures, HQ-SAM~\cite{HQSAM} supplements SAM by introducing a learnable high-quality output token, thereby enhancing its performance across diverse segmentation contexts.
Although both SAM and HQ-SAM excel in image segmentation, their direct application to video segmentation remains limited due to the need for point, box, or polygonal proposals to generate target mask results. 
Hence, we utilize gtboxes as prompts for HQ-SAM model to facilitate the generation of pixel-level predictions for the target object.
\begin{figure*}[]
	\centering
	\scalebox{0.23}{\includegraphics{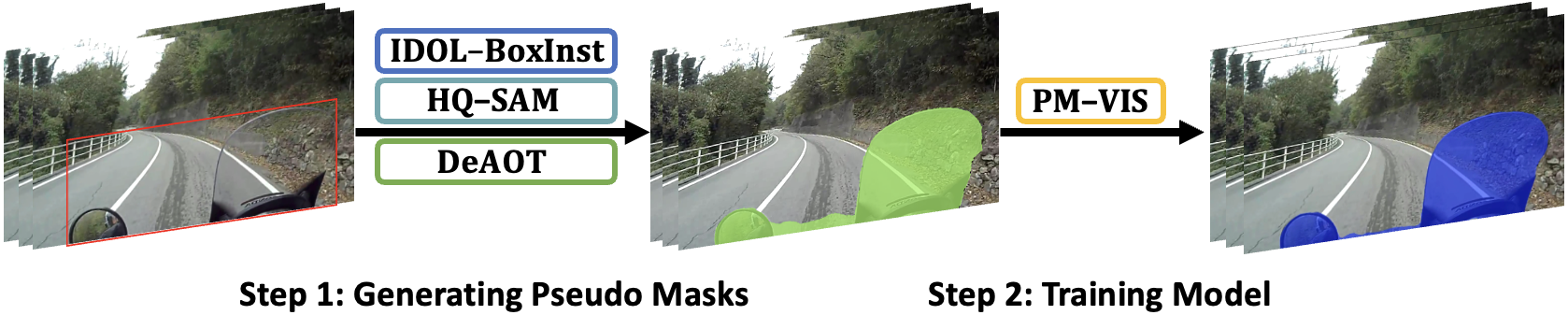}}
	\caption{
 Illustration of high-performance box-supervised VIS. Given a dataset with box annotations, our method consists of two steps. 
 }
	\label{fig:paperpipeline} 
\end{figure*}
\begin{figure*}[]
	\centering
	\scalebox{0.2}{\includegraphics{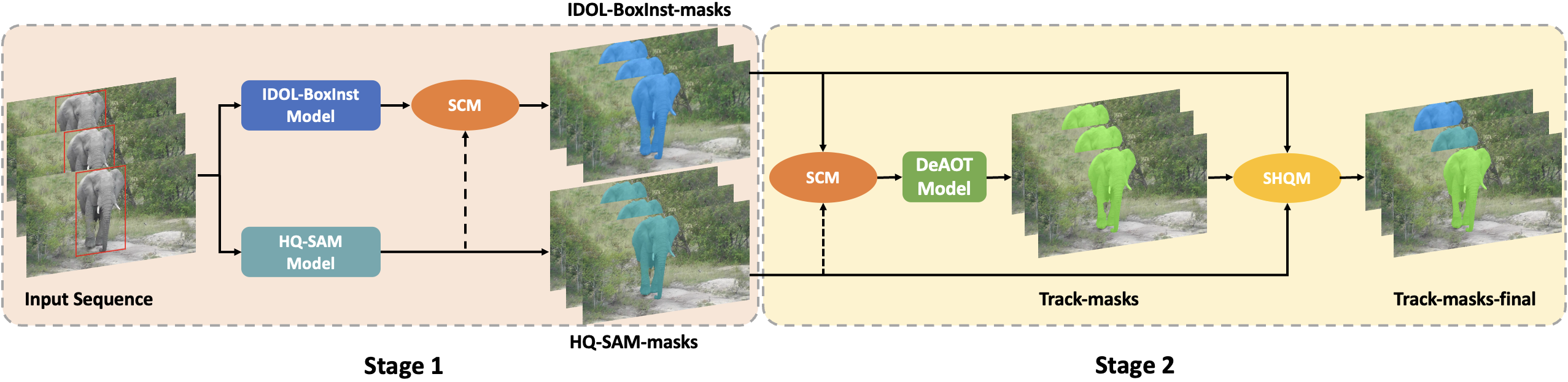}}
	\caption{
 Pipeline for high-quality pseudo masks generation. Given a video sequence with box annotations, our pipeline consists of two stages. In Stage 1, we employ the HQ-SAM model and the IDOL-BoxInst model to generate pixel-level predictions for the target objects, resulting in HQ-SAM-masks and IDOL-BoxInst-masks, respectively. Stage 2 involves selecting appropriate keyframes from the IDOL-BoxInst-masks to initialize the DeAOT model, resulting in higher-quality pixel-level predictions, represented as Track-masks. Track-masks-final is derived by selecting higher-quality pseudo masks from HQ-SAM-masks, IDOL-Box-masks and Track-masks using the SHQM method. Notably, dashed lines indicate auxiliary usage where HQ-SAM-masks is not directly utilized in subsequent applications. Conversely, solid lines represent direct usage. Two auxiliary mechanisms, SCM and SHQM, are employed to assist in selecting high-quality candidates for pseudo masks.
 }
	\label{fig:illpseudomasks} 
\end{figure*}

\section{METHODOLOGY}
Firstly, we present the pipelines for high-performance box-supervised VIS and the acquisition of high-quality pseudo masks in Section~\ref{sec:Pipeline}. Then, in Section~\ref{sec:Algorithm}, we extend the pixel-supervised VIS method IDOL to the box-supervised VIS (IDOL-BoxInst) baseline and introduce the box-supervised VIS (PM-VIS). Next, in Sections~\ref{sec:Pseudomaks} and~\ref{sec:methodforhigherqualitytrainset}, we describe several approaches for generating pseudo masks based on box supervision, and strategies for improving the quality of pseudo masks, respectively. Lastly, in Section~\ref{sec:FilterMethods}, we describe two methods for filtering ground-truth data with the help of Track-masks-final.

\subsection{Pipelines}	
\label{sec:Pipeline}
\subsubsection{High-performance box-supervised VIS}
Fig.~\ref{fig:paperpipeline} illustrates the workflow of high-performance box-supervised VIS, involving a two-step strategy. In Step 1, we obtain a training set with high-quality pseudo masks by leveraging three segmentation models. In Step 2, the PM-VIS model is trained using a combination of box annotations and the aforementioned pseudo masks. Through these procedures, our PM-VIS model achieves SOTA performance.
\subsubsection{High-quality pseudo masks}
The pipeline in Fig.~\ref{fig:illpseudomasks} summarizes our method concisely for generating high-quality pseudo masks through two stages.
In Stage~1, we employ IDOL-BoxInst and HQ-SAM to process the training dataset under box supervision, yielding two types of pseudo masks: IDOL-BoxInst-masks and HQ-SAM-masks. 
We establish correspondence between IDOL-BoxInst-masks and gtboxes by calculating SCM scores with the assistance of HQ-SAM-masks and gtboxes. Additionally, the proposed post-processing optimization strategy, DOOB, is applied to both types of pseudo masks obtained in Stage 1, resulting in refined pseudo masks.
In Stage 2, we employ VOS model DeAOT to track instances in the video and obtain high-quality pseudo masks. 
At this stage, there are two points that we need to note specifically.
Firstly, we initialize the model using pseudo masks from keyframes and subsequently track the pixel-level prediction information for the targets throughout the entire video.
Among these, keyframes are selected from the positions of certain instance pseudo masks in IDOL-BoxInst-masks determined by the SCM scores calculated between IDOL-BoxInst-masks, HQ-SAM-masks, and gtboxes.
Secondly, to further enhance pseudo mask quality, we calculate the confidence scores (SHQM) for HQ-SAM-masks, IDOL-BoxInst-masks, and Track-masks. Then, based on these scores, we choose the higher-quality pseudo masks from HQ-SAM-masks or IDOL-BoxInst-masks to substitute for the instance pseudo masks in the original Track-masks.

\subsection{Modules}	
\label{sec:Algorithm}

\subsubsection{Projection Loss}
The Projection Loss~\cite{BoxInst} utilizes gtbox annotation to supervise the horizontal and vertical projections of the predicted mask, enhancing the alignment between the predicted mask and the gtbox region.
The Projection Loss is formulated as:
\begin{equation}
  L_{proj} = L_{proj\_x}+L_{proj\_y},
  \label{eq:important}
\end{equation}
\begin{equation}
  L_{proj\_p} = L_{dice}(\max_p(m),\max_p( b )), p\in\{x,y\},
\end{equation}
where $L_{dice}$ represents the Dice loss defined in BoxInst~\cite{BoxInst}, $p$ denotes either the horizontal or vertical axis, $\max$ is the max operations along with each axis, $m$ corresponds to the predicted mask, and $b$ signifies the ground-truth box mask. 

\subsubsection{Pairwise Affinity Loss}

The Pairwise Affinity Loss~\cite{BoxInst} supervises pixel-level mask predictions without pixel-wise annotations by considering the color similarity between pixels. Specifically, for two pixels at coordinates $(i, j)$ and $(l, k)$, their color similarity is denoted as $S_{e} = \exp\left(-\frac{\parallel c_{i,j} - c_{l,k}\parallel}{\theta}\right)$.
Here, $c_{i,j}$ and $c_{l,k}$ represent color vectors in the LAB color space, with $\theta$ as a hyper-parameter (default as 2).
The color similarity threshold $\tau$ is set at 0.2, labeling the edges with similarity exceeding this threshold as 1. The network predicts $m_{i,j}\in(0, 1) $ as the probability of pixel $(i, j)$ being part of the foreground. Consequently, the probability $y_e=1$ for pairwise mask affinity is formulated as $P(y_e =1 )= m_{i,j} \cdot m_{l,k} +(1-m_{i,j})\cdot(1-m_{l,k})$.
The Pairwise Affinity Loss is defined as:
\begin{equation}
    L_{pair} = -\frac{1}{N}\sum \limits_{e\in E_{in}}\mathbbm{1}_{\{S_{e}>\tau \}} \log P(y_e=1),
\end{equation}
where $E_{in}$ is the set of edges containing at least one pixel in the box, and $N$ is the number of edges in $E_{in}$. The indicator function $\mathbbm{1}_{\{S_e>\tau\}}=1$ if $S_e>\tau$, and $0$ otherwise.

\subsubsection{Pseudo mask supervision loss}
By substituting the gtmasks with the pseudo masks, the pixel-wise mask supervision losses in the pixel-supervised VIS methods can be extended to the pseudo masks supervision losses, comprising the Focal Loss~\cite{FocalLoss} and the Dice Loss~\cite{DiceLoss}.

\subsubsection{IDOL-BoxInst and PM-VIS}
The IDOL-BoxInst algorithm is a box-supervised VIS method that combines the principles of box-supervised losses~\cite{BoxInst} with the IDOL algorithm~\cite{IDOL}, effectively eliminating the need for mask annotations and related losses~\cite{DiceLoss,FocalLoss}. 
PM-VIS is trained using a combination of spatio-temporal losses, including BoxInstLoss~\cite{BoxInst}, which comprises Projection Loss and Pairwise Affinity Loss from BoxInst~\cite{BoxInst}, as well as pixel-wise mask supervision losses (Focal loss~\cite{FocalLoss} and Dice loss~\cite{DiceLoss}), referred to as MaskLoss.
PM-VIS adopts a weight configuration similar to that of IDOL and BoxInst, setting the weight (W1) for BoxInstLoss to 1 and the weight (W2) for MaskLoss to 0.5.
\subsection{Pseudo masks}
\label{sec:Pseudomaks}
Current VIS training datasets (YTVIS2019~\cite{VIS}, YTVIS2021~\cite{YTVIS2021}, and OVIS~\cite{OVIS}) for box-supervised VIS include information such as target categories and bounding boxes, yet lacks pixel-level annotations for the designated targets. In the context of VIS, a model trained only with annotations of target bounding boxes typically exhibits diminished recognition capabilities compared to a model trained with mask annotations. 
Hence, by leveraging three models (HQ-SAM, IDOL-BoxInst, and DeAOT), we generate three types of pseudo masks (HQ-SAM-masks, IDOL-BoxInst-masks, and Track-masks) based on box annotations, further enhancing our VIS performance.
\subsubsection{HQ-SAM-masks}	
The HQ-SAM-masks, obtained by processing video frames based on the proposals of gtboxes using the HQ-SAM~\cite{HQSAM} model, serves as pseudo mask annotations for the training set. In this work, we train the PM-VIS model using pseudo-mask supervision on the mentioned datasets. We observe that the VIS performance is better with the use of pseudo masks compared to without using them. However, due to the limited quality of HQ-SAM-masks, the improvement is constrained.
\subsubsection{IDOL-BoxInst-masks}	
\begin{figure}[h]
	\centering
	\scalebox{0.13}{\includegraphics{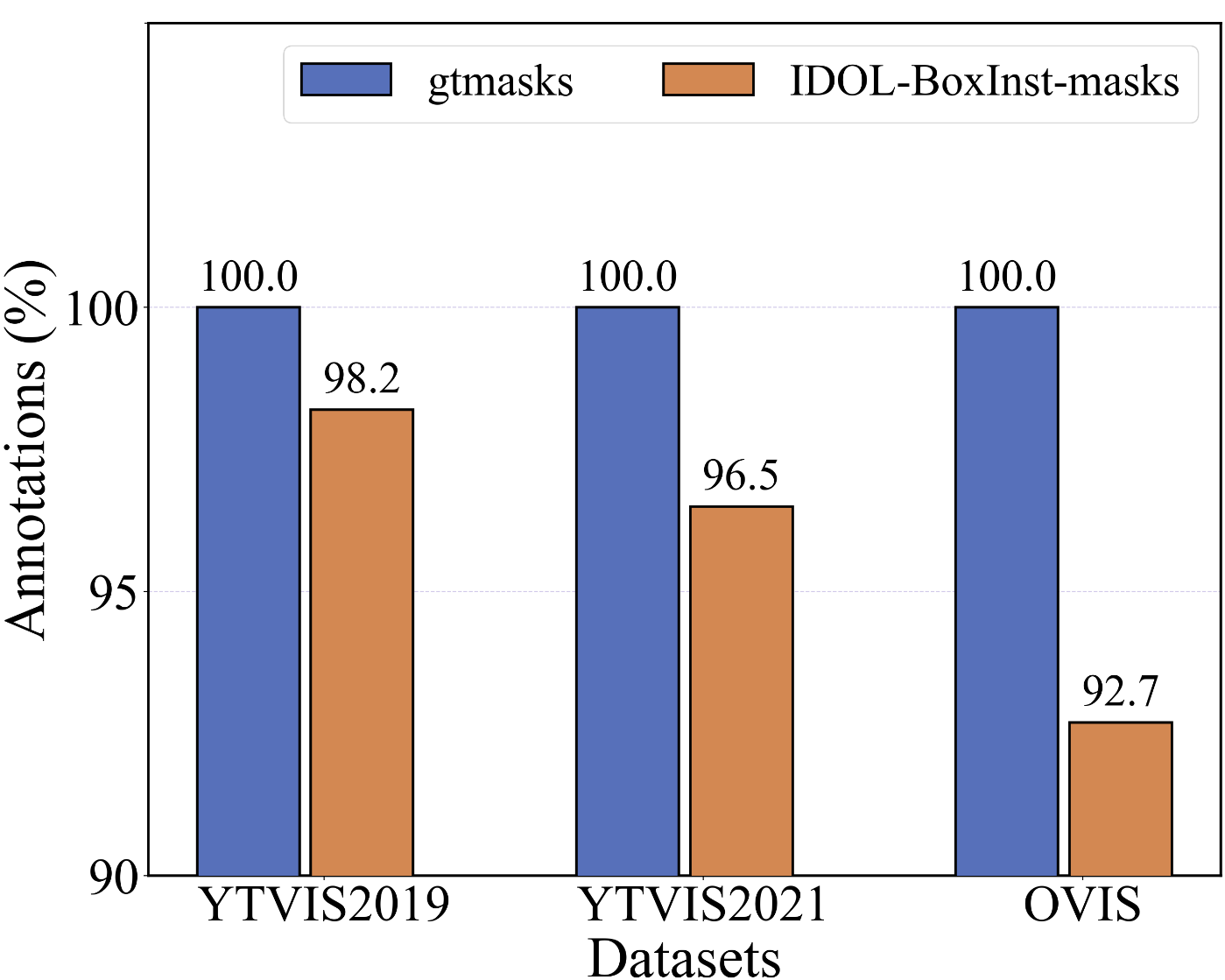}}
	\caption{Comparison of the annotated instances in IDOL-BoxInst-masks (or Track-masks) with the annotated instances in the ground-truth data. The horizontal and vertical axes represent the datasets and the percentage of instance annotations within each datasets, respectively.}
	\label{fig:missing-data-percent-fig} 
\end{figure}

As illustrated in Fig.~\ref{fig:illpseudomasks} (Stage 1), we generate another type of pseudo masks, namely IDOL-BoxInst-masks, in addition to the pseudo masks generated by the HQ-SAM model. The IDOL-BoxInst model, trained without mask annotations, generates these pseudo masks by predicting the targets in the training set.
Nonetheless, the current pseudo masks obtained from this process exhibit several concerns: \textbf{(i)} the predicted masks might exhibit suboptimal quality and inherent errors, \textbf{(ii)} the labels corresponding to the predicted or tracked masks could be inaccurate, and \textbf{(iii)} multiple distinct masks may be independently predicted for the same target, resulting in pixel-level ambiguity and potentially compromising experimental precision. To address these challenges, we introduce an innovative predictive mask allocation and correction strategy called \textbf{SCM} (as described in Section~\ref{sec:SCM}) that aims at using HQ-SAM-masks to aid in establishing a coherent relationship between the predicted masks and gtboxes for certain instances.
Additionally, we conduct an annotation quantity analysis and find that the number of instance annotations for IDOL-BoxInst-masks is relatively lower compared to the ground-truth datasets.
Specifically, as shown in Fig.~\ref{fig:missing-data-percent-fig}, the quantity of IDOL-BoxInst-masks obtained in our experiments is found to be 1.8\%, 3.5\%, and 7.3\% lower than the number of instances annotated in the ground-truth for YTVIS2019, YTVIS2021, and OVIS, respectively.

\subsubsection{Track-masks}	
As shown in Fig.~\ref{fig:illpseudomasks} (Stage 2), Track-masks is a collection of pseudo masks obtained by initializing the semi-supervised VOS model DeAOT with instance mask annotations from IDOL-BoxInst-masks and tracking the instances throughout the entire video.
Typically, semi-supervised VOS models require information about the target in the first frame of the video during prediction.
However, for our task, the initial frame instance pseudo mask may not be optimal. To overcome this limitation, we propose a mechanism to initiate tracking from various positions in the video, progressing both forwards and backwards until reaching the start and end frames, respectively. Subsequently, the tracked results are amalgamated, and these starting positions are termed keyframes.
Specifically, our procedure is as follows: firstly, we employ the SCM method to create a collection of high-quality keyframes from the IDOL-BoxInst-masks. 
Secondly, we obtain the pseudo mask collection, Track-masks, by feeding the annotations from these keyframes into the DeAOT model for prediction.
Finally, we employ two methods to further enhance the quality of Track-masks. These methods involve expanding the keyframes for each instance from one to $len(video)/K$ and selecting high-quality pseudo masks for optimizing Track-masks from both IDOL-Boxlnst-masks and HQ-SAM-masks using the SHQM method.
Here, $len(video)$ denotes the video length, and we configure $K$ as a hyper-parameter, setting it to be 10 for YTVIS2019/YTVIS2021 and 100 for OVIS.
\subsection{Strategies}
\label{sec:methodforhigherqualitytrainset}
As shown in Fig.~\ref{fig:illpseudomasks} and Fig.~\ref{fig:beyondbboxmaskOverlap}, to acquire high-quality pseudo masks, we propose two confidence assessment methods, SCM and SHQM, along with a post-processing technique, DOOB. SCM is employed to identify the most suitable gtboxes for the targets in the frames, assisted by HQ-SAM-masks, both for refining the masks predicted by the IDOL-BoxInst model and for selecting a specific number of keyframes. SHQM serves to choose superior-quality pseudo masks from HQ-SAM-masks, IDOL-BoxInst-masks, and Track-masks, thereby reassembling them into the highest-quality pseudo mask collection, Track-masks-final. Meanwhile, DOOB functions as a post-processing mechanism to rectify erroneous areas within HQ-SAM-masks and IDOL-BoxInst-masks.

\subsubsection{SCM}	
\label{sec:SCM}
We present a methodology to quantify the correlation between pseudo masks, as shown in Eq.~(\ref{eq:SCM}). Let $m_1$ and $m_2$ denote two distinct types of pseudo masks. We calculate the IoU between these masks and also compute the IoU~\cite{GIoU} between the bounding boxes of these masks and the gtbox. The combination of these three factors generates a score, denoted as the \textbf{S}core of \textbf{C}orresponding \textbf{M}ask (\textbf{SCM}), signifying the relationship between the two pseudo masks. 
\begin{equation}
\label{eq:SCM}
\begin{aligned}
	SCM=IOU(m_1,m_2)\cdot H_1\cdot H_2 ,
\end{aligned}
\end{equation}
\begin{equation}
	 H_i=IOU(Box(m_i),gtbox) ,i\in\{1,2\},
\end{equation}
where $Box(m)$ represents the bounding box of a given pseudo mask, $IOU(\cdot, \cdot)$ represents the calculation of IoU, while $gtbox$ denotes the ground-truth bounding box of the target.
\subsubsection{SHQM}	
\label{sec:SHQM}
We introduce another technique called \textbf{SHQM} for \textbf{S}electing the \textbf{H}igh-\textbf{Q}uality \textbf{M}asks among the three variants: HQ-SAM-masks, IDOL-BoxInst-masks, and Track-masks. 
This method establishes a composite connection between each pseudo mask  and the gtbox, generating confidence scores $S_{ij}$ for every pseudo mask. 
The pseudo mask attaining the highest score, labeled as $m_{\text{z}}$, indicates the superior-quality pseudo mask.
\begin{equation}
	 z=\underset{i}{\text{argmax}} \, A_i,i\in\{1,2,3\} ,
\end{equation}
\begin{equation}
	 A_i= \sum\limits_{j}S_{ij},i \neq j,i,j\in\{1,2,3\} ,
\end{equation}
\begin{equation}
\begin{aligned}
	 {S_{ij}} = SCM(m_i,m_j),i \neq j,i,j\in\{1,2,3\} .
\end{aligned}
\end{equation}
This relationship is established through the assessment of the SCM denoted as $S_{ij}$ between different pairs of pseudo masks $m_i$ and $m_j$, where $i$ and $j$ are members of the set $\{1,2,3\}$ and are distinct from each other. $S_{ij}$ is equal to $S_{ji}$ here.

\subsubsection{DOOB}
\begin{figure}[h]
	\centering
	\scalebox{0.13}{\includegraphics{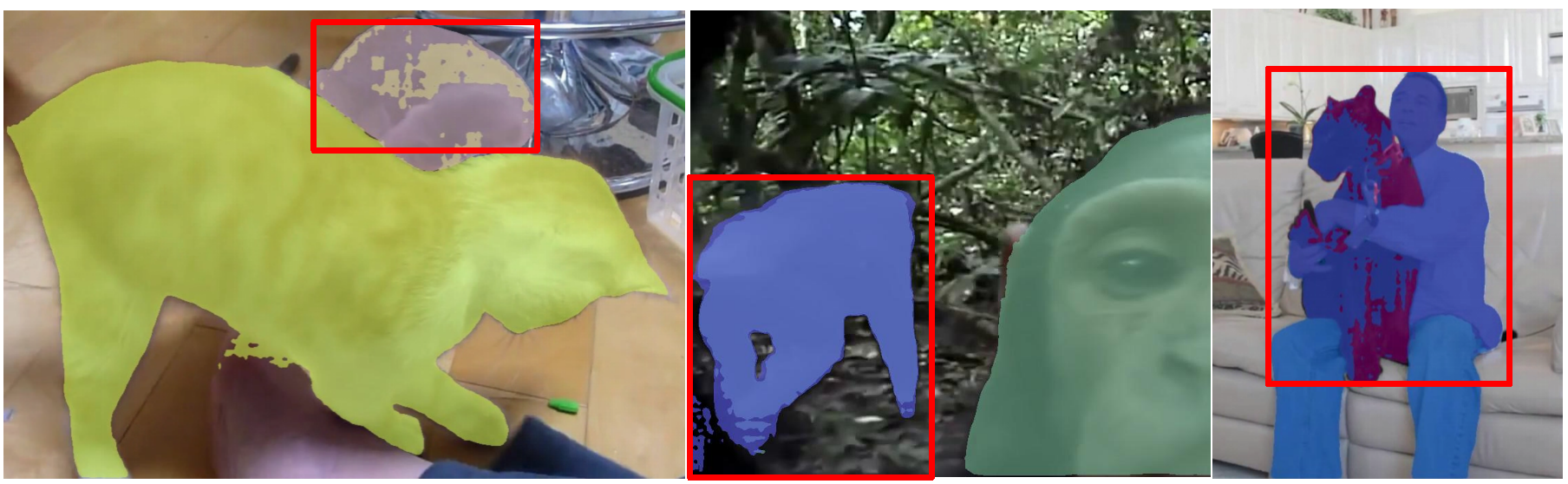}}
	\caption{Some incorrect mask predictions. The erroneous regions, which either overlap or extend beyond the boundaries, are highlighted within the red boxes. In the first two images, there are overlaps between target mask annotations and extensions beyond the respective gtbox for other targets. In the last image, two target mask annotations overlap.}
	\label{fig:beyondbboxmaskOverlap} 
\end{figure}

After analyzing the pseudo mask datasets, two significant issues in the predicted masks have become evident: \textbf{(i)}~overlapping occurrences between distinct predicted masks within the same frame, and \textbf{(ii)} the expansion of predicted masks beyond the boundaries of the corresponding gtbox, as illustrated in Fig.~\ref{fig:beyondbboxmaskOverlap}. To address these concerns and enhance the accuracy of the pseudo mask training dataset, we introduce an optimization approach named \textbf{DOOB}, which involves \textbf{D}eleting the \textbf{O}verlapping and \textbf{O}ut-of-\textbf{B}oundary sections. Specifically, the strategy ascertains ownership of overlapping regions by computing the bounding rectangle of the overlapping mask and gauging the IoU~\cite{GIoU} value with the associated gtbox. Subsequently, it eliminates mask annotations for the current target that exceed a reasonable range determined by the gtbox.


\subsection{Filter Methods}
\label{sec:FilterMethods}

Evidently, there is a noticeable gap between the current pseudo mask data and the ground-truth data. Through analysis, it has been observed that, in addition to the quality differences in individual instance mask annotations, there is also an issue of instance mask annotations absence, as demonstrated in Fig.~\ref{fig:missing-data-percent-fig}.
This outcome can be attributed to two primary factors. Firstly, the limited recognition capability of the IDOL-BoxInst model makes it unable to fully identify all instances in the video. Secondly, in the process of assigning ownership through IoU calculations between IDOL-BoxInst-masks and predicted masks, instances may be wrongly allocated to other instances due to suboptimal quality in either the predicted masks or HQ-SAM-masks.
Therefore, we surmise that these missing annotations might also pose challenges for the algorithm, potentially leading to misleading guidance.
Following experimental results demonstrate that these missing instances do have a negative effect on the performance, so we exclude the information about these missing instances from Track-masks.
Besides, during the training of the fully supervised model using ground-truth data, we exclude the mapping annotations for these missing instances, and this operation improves Mask Average Precision (AP) of our model.

An undeniable fact is that the quality of pseudo masks exhibits noticeable deficiencies compared to ground truth data, and in weakly supervised tasks, these shortcomings undoubtedly constrain algorithm performance. However, in a fully supervised task, these deficiencies might play a facilitating role. Therefore, to fully harness the potential of pseudo masks, we propose utilizing them to filter the ground truth data.
\subsubsection{Missing-Data}
\begin{figure*}[t]
	\centering
	\scalebox{0.344}{\includegraphics{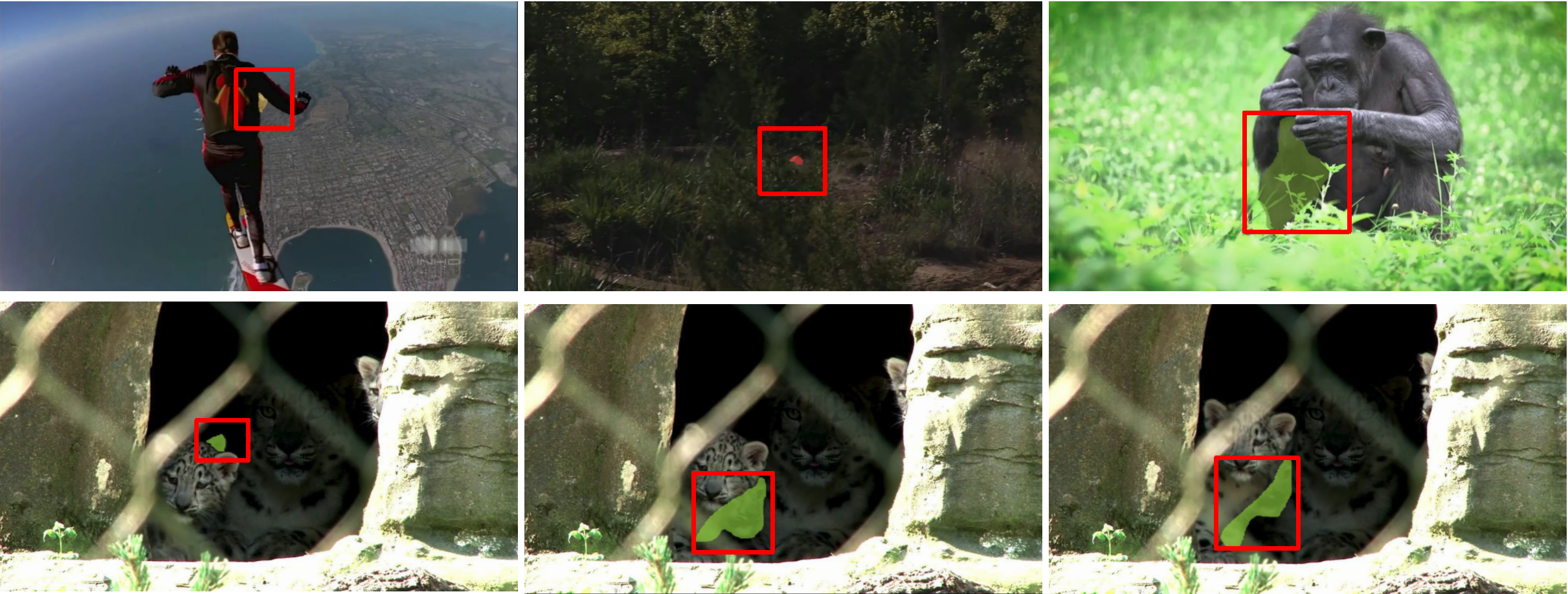}}
	\caption{Visualizing ground-truth data that results from mapping and subsequently removing missing instance masks in IDOL-BoxInst-masks (or Track-masks).}
	\label{fig:gtdifference} 
\end{figure*}
During the experiments, it becomes apparent that the number of Track-masks is lower than that of the ground-truth data. Upon visualization, as depicted in Fig.~\ref{fig:gtdifference}, we identify several issues with these missing data, including indistinct instance features, low discriminability of instances from surrounding objects or the environment, and severe occlusions. 
Thus, we propose a filtering method named \textbf{Missing-Data} to remove unnecessary \textbf{Missing Data} from the ground-truth. Specifically, this method leverages our proposed pseudo mask set, Track-masks-final, as a reference to eliminate corresponding discrepancies in the ground-truth data, resulting in a ground-truth dataset of the same size as Track-masks-final.

\subsubsection{RIA}
\begin{figure*}[!t]
	\centering
	\scalebox{0.17}{\includegraphics{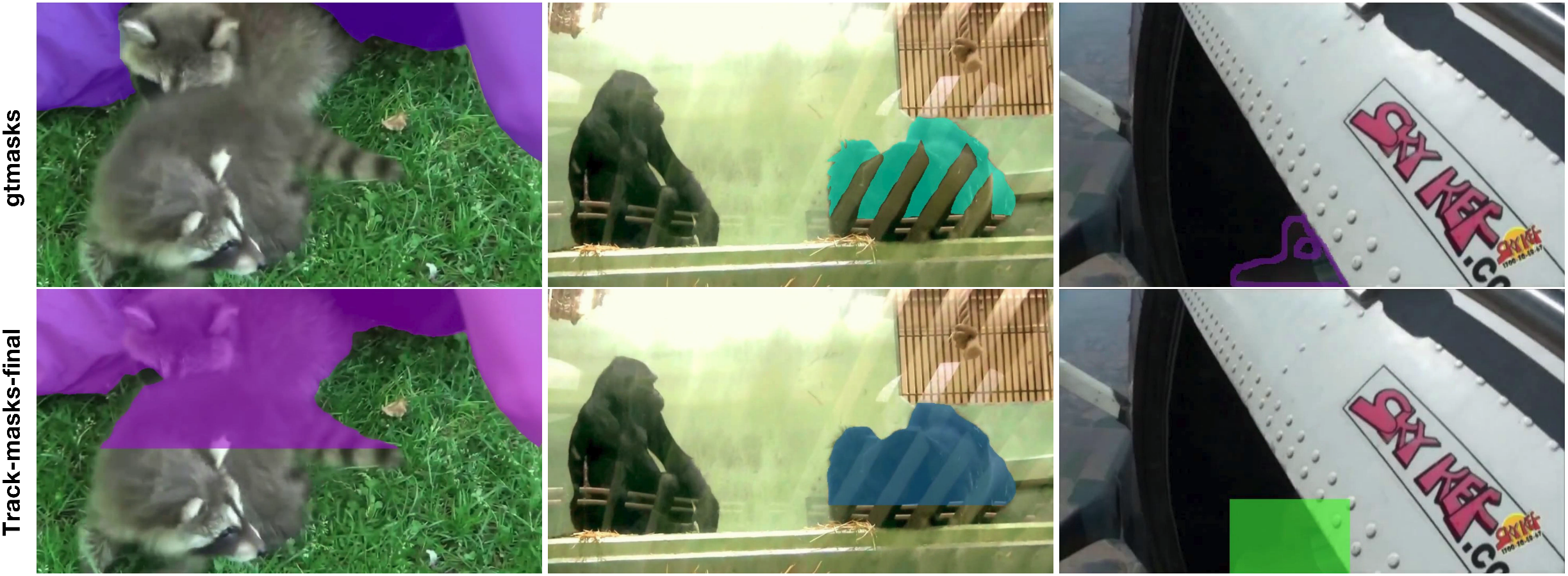}}
	\caption{Visualizing the data removed from the ground-truth data using the RIA method. The first row represents the data to be excluded from gtmasks, while the second row represents the corresponding data from the auxiliary dataset Track-masks-final.}
	\label{fig:maskioubelow06} 
\end{figure*}
Our pseudo masks have substantially improved the performance of weakly supervised VIS. However, it's inevitable that the pseudo mask data we obtained, Track-masks-final, still exhibit disparities in instance segmentation quality compared to the ground-truth, as shown in Fig.~\ref{fig:maskioubelow06}.
Specifically, Track-masks-final tends to struggle with small, complex, and less distinct instances. While this issue poses a challenge in weakly supervised settings, it can be reframed as an optimization problem when considered under fully supervised conditions. Therefore, we propose a method called \textbf{RIA}. This method involves calculating the mask IoU between gtmasks and Track-masks-final data to map and \textbf{R}emove \textbf{I}nstance \textbf{A}nnotations in the ground-truth with relatively low mask IoU values. 
As illustrated in Fig.~\ref{fig:maskioubelow06} (gtmasks), datasets characterized by lower IoU values frequently encounter challenges, including inconspicuous features and severe occlusions. 
Consequently, this refinement and optimization process enhances the quality of the ground-truth data, yielding a more compact yet higher-quality dataset.
\section{Experiments}
\subsection{Datasets and Metrics}
The proposed approach in this paper involves datasets for two distinct tasks: YTVIS2019~\cite{VIS}, YTVIS2021~\cite{YTVIS2021}, and OVIS~\cite{OVIS} for VIS, as well as YTVOS18~\cite{Youtube-vos}, YTVOS19~\cite{vos2019} and DAVIS~\cite{davis} for VOS.
\subsubsection{VIS}
YTVIS2019, a subset of YTVOS18/19, comprises 2883 videos, 40 categories, 4483 objects, and over 130k annotations. YTVIS2021, an improved version of YTVIS2019, has expanded its dataset, refined the 40-category label set, and thus partially overlaps with YTVOS18/19. 
OVIS, one of the most challenging datasets in the VIS domain, presents significant hurdles. Instances within OVIS exhibit severe occlusion, intricate motion patterns, and rapid deformations. The dataset consists of 607 training videos and 140 validation videos, with notably extended video durations averaging 12.77 seconds.
Regarding VIS evaluation, we employ standard metrics such as AP, AP50 (Average Precision at 50\% IoU), AP75 (Average Precision at 75\% IoU), AR1 (Average Recall at 1), and AR10 (Average Recall at 10).
\subsubsection{VOS}
YTVOS18/19 and YTVIS2019/2021 represent two distinct datasets utilized for VOS and VIS tasks, respectively. However, in terms of data sources, all data from YTVIS2019 and some data from YTVIS2021 belong to YTVOS18/19. In this paper, we use DeAOT as a tracking model to obtain pseudo masks. To ensure the credibility of our task and avoid the impact of overlapping data, we exclude the overlapping data when retraining DeAOT using data from YTVOS18/19, as shown in Table~\ref{table:noOverlapVosdata}. The overlapping data between YTVIS2019/2021 and YTVOS18/19 accounts for 64.5\% of the data in YTVOS18/19, while the non-overlapping data accounts for 35.5\%. As for VOS evaluation, we adopt the evaluation approach from the DeAOT algorithm, including metrics like the region similarity $J$, the contour accuracy $F$, and their mean value (referred to as $J\&F$).

\begin{table}[h]
    \small
	\centering
	\renewcommand\arraystretch{1.3}
	\setlength\tabcolsep{5pt} 
	\caption{The overlapping quantity of YTVIS2019, YTVIS2021, and YTVOS18/19. The diagonal elements represent the count of videos in the training set of the dataset. Off-diagonal data, such as 2191/73.4\%, indicate an overlap of 2191 videos between the training sets of YTVIS2019 and YTVIS2021, accounting for 73.4\% of the YTVIS2021. There are 2238 videos that overlap between the YTVIS2019/2021 and YTVOS18/19 datasets, accounting for 64.5\% of the YTVOS18/19. }
	\label{table:noOverlapVosdata}
\begin{tabular}{l|l|l|l}
\hline
           & YTVIS2019        & YTVIS2021       & YTVOS18/19     \\ \hline\hline
YTVIS19    & 2238 / 100.0\% & -             & -              \\ \hline
YTVIS21    & 2191 / 73.4\%    & 2985 / 100.0\%     & -              \\ \hline
YTVOS18/19 & 2238 / 64.5\%  & 2191 / 63.1\% & 3471 / 100.0\% \\ \hline
\end{tabular}
\end{table}

\subsection{Implementation Details}
For VOS, we fully follow the configuration in the previous DeAOT~\cite{DeAOT} algorithm. For VIS, unless otherwise stated, we employ the same hyper-parameters as IDOL~\cite{IDOL}, which is implemented on top of Detectron2~\cite{detectron2}. 
All our VIS models have been pre-trained on the COCO~\cite{COCO} image instance segmentation dataset with pixel-wise annotations.

\textbf{Model Settings.} Apart from integrating the two box-supervised segmentation loss terms from BoxInst~\cite{BoxInst}, our proposed algorithm only applies necessary adjustments to the weights of mask losses~\cite{DiceLoss,FocalLoss} from IDOL.
The training and testing details are kept as similar as possible to the original IDOL~\cite{IDOL}.
Unless specified otherwise, we use ResNet-50~\cite{ResNet-50} as the backbone.

\textbf{Training.} Our training process has been meticulously aligned with IDOL~\cite{IDOL}, covering aspects such as the optimizer, weight decay, and the utilization of the COCO pre-trained model provided by IDOL.
For YTVIS2019/2021 datasets, we perform downscaling and random cropping on input frames to ensure that the longest side measures no more than 768 pixels.
For OVIS, when using the ResNet-50 backbone, we keep the same configuration as training IDOL. When using the Swin Large~\cite{Swin-L} (Swin-L) backbone, we resize input images to ensure that the shortest side falls within a range of 480 to 736 pixels, and the longest side remains below 736 pixels.
Training is conducted on 4 RTX3090 GPUs, each with 24GB of RAM, allocating at least 2 frames per GPU.

\textbf{Inference.} During inference, the input frames are downscaled to 360 pixels for YTVIS2019/2021 following IDOL, and 720 pixels for OVIS, given its videos have a higher resolution. 
Regarding hyper-parameters, we default to following the specifications outlined in IDOL.

\subsection{Ablation study}
In our ablation studies on the YTVIS2019 validation set, ResNet-50 is employed as the backbone architecture.
\begin{table}[h]
    \small
	\centering
	\renewcommand\arraystretch{1.3}
	\setlength\tabcolsep{10pt} 
	\caption{Effect of the weight W1 of BoxInstLoss on YTVIS2019 val. Setting the weight of MaskLoss W2=0.5.}
	\label{table:masklossW1}
	\begin{tabular}{l|l|l|l|l|l}
  \hline
  W1  & AP            & AP50          & AP75          & AR1           & AR10          \\ \hline\hline
  0   & 46.6          & 72.5          & 50.4          & 43.7          & 53.0          \\
  0.5 & 47.3          & 71.9          & 51.0          & 44.2          & 54.9          \\
  1   & \textbf{48.7} & \textbf{73.4} & \textbf{52.4} & \textbf{45.2} & 55.3          \\
  1.5 & 46.9          & 71.8          & 50.3          & 44.5          & \textbf{55.9}\\ \hline
  \end{tabular}
\end{table}

\begin{table}[h]
    \small
	\centering
	\renewcommand\arraystretch{1.3}
	\setlength\tabcolsep{10pt} 
	\caption{Effect of the weight W2 of MaskLoss on YTVIS2019 val. Setting the weight of BoxInstLoss W1=1.}
	\label{table:masklossW2}
	\begin{tabular}{l|l|l|l|l|l}
    \hline
    W2                     & AP            & AP50          & AP75          & AR1           & AR10          \\ \hline\hline
    0                      & 44.8          & 71.2          & 48.3          & 41.8          & 52.9          \\
    0.3                    & 47.4          & 72.2          & 51.4          & 43.9          & 54.9          \\
    0.5                    & \textbf{48.7} & \textbf{73.4} & \textbf{52.4} & 45.2          & 55.3          \\
    0.8                    & 47.9          & 72.8          & 51.6          & \textbf{45.5} & \textbf{56.6} \\
    \multicolumn{1}{l|}{1} & 47.0          & 72.1          & 50.5          & 44.7          & 55.2         \\ \hline
    \end{tabular}
\end{table}

\begin{table}[h]
    \small
	\centering
	\renewcommand\arraystretch{1.3}
	\setlength\tabcolsep{4pt} 
	\caption{Effect of pseudo masks on YTVIS2019 val. IBM: IDOL-BoxInst-masks. The term ``Track-masks + multiframe + SHQM" refers to Track-masks-final.}
	\label{table:vspseudomasks}
  \begin{tabular}{l|l|l|l|l|l}
    \hline
    Method                       & AP   & AP50 & AP75 & AR1  & AR10 \\ \hline\hline
    HQ-SAM-masks                       & 46.8  & 72.3 & 49.9 & 45.3 & 55.3 \\
    +DOOB                & 46.8  & 73.4 & 51.5 & 43.5 & 53.8 \\ \hline
    IBM+SCM                        & 45.7 & 72.8 & 47.9  &43.6 & 53.9 \\
    IBM+SCM+DOOB           & 46.9 &73.2&49.8&45.3&55.4  \\ \hline
    Track-masks                        & 48.2 & 72.8 & 51.7 & 44.8 &\textbf{55.9}  \\
    +SHQM              & 47.4 & 72.9 & 51.6 & 44.2 &54.8  \\
    +multiframe+SHQM  & \textbf{48.7} & \textbf{73.4} & \textbf{52.4} & \textbf{45.2} & 55.3\\ \hline
    \end{tabular}
\end{table}

\begin{table}[h]
    \small
	\centering
	\renewcommand\arraystretch{1.3}
	\setlength\tabcolsep{4pt} 
	\caption{Comparison of the effects of IDOL-BoxInst-masks and Track-masks-final in RIA. }
	\label{table:twopseudomasksmaskIOUgTvs}
\begin{tabular}{l|l|l|l}
\hline
Method                  & Sup.                   & Pseudo masks       & AP \\ \hline\hline
\multirow{2}{*}{PM-VIS} & \multirow{2}{*}{Pixel} & IDOL-BoxInst-masks & 48.7\\
                        &                        & Track-masks-final       & 50.0 \\ \hline
\end{tabular}
\end{table}

\begin{table}[h]
    \small
	\centering
	\renewcommand\arraystretch{1.3}
	\setlength\tabcolsep{10pt} 
	\caption{The impact of BoxInstLoss on fully supervised VIS. The result with superscript ``\dag " indicates the MaskLoss weight is set to 0.5.}
	\label{table:BoxInstLossforfullysup}
\begin{tabular}{l|l|c|l}
\hline
Method & Sup. & BoxInstLoss & AP            \\ \hline\hline
IDOL\textsuperscript{\dag}  & \multirow{2}{*}{Pixel}    & \ding{55} & 47.7          \\
PM-VIS &                           & \checkmark  & \textbf{50.0} \\ \hline
\end{tabular}
\end{table}

\begin{table}[h]
    \small
	\centering
	\renewcommand\arraystretch{1.3}
	\setlength\tabcolsep{3pt} 
	\caption{The impact of two filtering methods, Missing-Data and RIA, on YTVIS2019 ground-truth data. }
	\label{table:missingdatamaskIOUgT}
\begin{tabular}{l|l|c|c|l}
\hline
Method                        & Sup.                   & Missing-Data     &  RIA    & AP            \\ \hline\hline
\multirow{2}{*}{IDOL-BoxInst} & \multirow{2}{*}{Box}   &  \ding{55}       &  \ding{55}    & 43.9          \\
                              &                        & \checkmark       &  \ding{55}    & 44.8          \\ \hline
\multirow{4}{*}{PM-VIS}       & \multirow{4}{*}{Pixel} &  \ding{55}       &  \ding{55}    & 49.1          \\
                              &                        & \checkmark       &  \ding{55}    & 49.3          \\
                              &                        & \checkmark       & \checkmark    & \textbf{50.0} \\ \cline{3-5} \hline
\end{tabular}
\end{table}

\begin{table}[h]
    \small
	\centering
	\renewcommand\arraystretch{1.3}
	\setlength\tabcolsep{10pt} 
	\caption{Effect of the RIA filtering method with varying mask IoU thresholds ($\tau$) on fully supervised VIS.}
	\label{table:maskIOUgT}
\begin{tabular}{l|l|l|l}
\hline
Method                  & Sup.                               &  $\tau$   & AP     \\ \hline\hline
\multirow{6}{*}{PM-VIS} & \multirow{6}{*}[0.5ex]{Pixel}         & 0       & 49.3 \\
                        &                                            & 0.5     & 49.7  \\
                        &                                            & 0.6     & \textbf{50.0}  \\
                        &                                            & 0.7     & 48.2   \\
                        &                                            & 0.8     & 47.2   \\ \hline
\end{tabular}
\end{table}

\begin{table*}[ht]
    \small
	\centering
	\renewcommand\arraystretch{1.3}
	\setlength\tabcolsep{2.3pt} 
	\caption{Quantitative performance comparison of pixel-supervised and box-supervised VIS methods with \textbf{ResNet-50} backbone on YTVIS2019, YTVIS2021 and OVIS. Best in \textbf{bold} and second \underline{underline}. The results with superscript ``\dag " represent our self-trained outcomes. }
	\label{table:sotaResNet50}
  \begin{tabular}{lllllllllllllllll}
  \hline
  \multirow{2}{*}{Method}           & \multirow{2}{*}{Sup.} & \multicolumn{5}{c}{YTVIS2019}      & \multicolumn{5}{c}{YTVIS2021}      & \multicolumn{5}{c}{OVIS}         \\
                                    &                       & AP   & AP50 & AP75 & AR1  & AR10 & AP   & AP50 & AP75 & AR1  & AR10 & AP   & AP50 & AP75 & AR1  & AR10 \\ \hline\hline
  Mask2Former~\cite{Mask2formerVIS}                       & Pixel                 & 47.8 & 69.2 & 52.7 & 46.2 & 56.6 & 40.6 & 60.9 & 41.8 & -    & -    & 17.3 & 37.3 & 15.1 & 10.5 & 23.5 \\
  SeqFormer~\cite{Seqformer}                         & Pixel                 & 47.4 & 69.8 & 51.8 & 45.5 & 54.8 & 40.5 & 62.4 & 43.7 & 36.1 & 48.1 & 15.1 & 31.9 & 13.8 & 10.4 & 27.1 \\
  IDOL~\cite{IDOL}                              & Pixel                 & 49.5 & \textbf{74.0} & 52.9 & \underline{47.7} & \underline{58.7} & 43.9 & 68.0 & 49.6 & 38.0 & 50.9 & \textbf{30.2} & \underline{51.3} & 30.0 & \textbf{15.0} & 37.5 \\
  IDOL\textsuperscript{\dag}~\cite{IDOL}       & Pixel                 & 49.0 & 72.2 & \textbf{54.8} & 46.4 & 57.3 & \underline{46.6} & \underline{70.2} & \underline{50.6} & 40.5 & \textbf{56.0} & 29.4 &50.6 & \textbf{30.9} & \underline{14.9} & \underline{37.7} \\
  VITA~\cite{Vita}                              & Pixel                   & \underline{49.8} & 72.6 & \underline{54.5} & \textbf{49.4} & \textbf{61.0} & 45.7 & 67.4 & 49.5 & \underline{40.9} & 53.6 & 19.6 & 41.2 & 17.4 & 11.7 & 26.0 \\ 
  PM-VIS & Pixel&\textbf{50.0}&\underline{73.6}&\underline{54.5}&46.5&57.1&\textbf{47.7}&\textbf{70.6}&\textbf{52.6}&\textbf{41.2}&\underline{55.8}&\underline{29.9}&\textbf{54.0}&\underline{30.7}&14.2&\textbf{38.0}\\ \hline
  MaskFreeVIS~\cite{maskfreeVIS}                   & Box                   & \underline{46.6} & \underline{72.5} & \underline{49.7} & \underline{44.9} & \textbf{55.7} & 40.9 & 65.8 & 43.3 & \underline{37.1} & \underline{50.5} & 15.7 & 35.1 & 13.1 & 10.1 & 20.4    \\
  IDOL-BoxInst                      & Box                   & 43.9 & 71.0 & 47.8 & 42.9 & 52.7 & \underline{41.8} & \underline{67.4} & \underline{43.5} & 36.5 & 50.3 & \underline{25.4} & \underline{47.4} & \underline{23.7} & \underline{12.9} & \underline{32.7} \\
  \textbf{PM-VIS}& Box& \textbf{48.7} & \textbf{73.4} & \textbf{52.4} & \textbf{45.2} & \underline{55.3} & \textbf{44.6} & \textbf{69.5} & \textbf{49.0} & \textbf{38.9} & \textbf{52.1}  &\textbf{27.8}&\textbf{48.5}&\textbf{27.4}&\textbf{13.6}&\textbf{36.0}\\ \hline
  \end{tabular}
\end{table*}

\begin{table*}[htbp]
    \small
	\centering
	\renewcommand\arraystretch{1.3}
	\setlength\tabcolsep{2.3pt} 
	\caption{Quantitative performance comparison of pixel-supervised and box-supervised VIS methods with \textbf{Swin-L} backbone on YTVIS2019, YTVIS2021 and OVIS. Best in \textbf{bold} and second \underline{underline}. The results with superscript ``\dag " represent our self-trained outcomes.}
	\label{table:sotaswinL}
  \begin{tabular}{lllllllllllllllll}
    \hline
    \multirow{2}{*}{Method} & \multirow{2}{*}{Sup.} & \multicolumn{5}{c}{YTVIS2019}      & \multicolumn{5}{c}{YTVIS2021}      & \multicolumn{5}{c}{OVIS}         \\
                                      &       & AP   & AP50 & AP75 & AR1  & AR10 & AP   & AP50 & AP75 & AR1  & AR10 & AP   & AP50 & AP75 & AR1 & AR10 \\ \hline\hline
    Mask2Former~\cite{Mask2formerVIS}  & Pixel & 60.4 & 84.4 & 67.0 & -    & -    & 52.6 & 76.4 & 57.2 & -    & -    & 26.4 & 50.2 & 26.9 & -   & -    \\
    SeqFormer~\cite{Seqformer}         & Pixel & 59.3 & 82.1 & 66.4 & 51.7 & 64.4 & 51.8 & 74.6 & 58.2 & 42.8 & 58.1 & -    & -    & -    & -    & -    \\
    IDOL~\cite{IDOL}                   & Pixel & \textbf{64.3} & \textbf{87.5} & \textbf{71.0} & \textbf{55.6} & \textbf{69.1} & 56.1 & 80.8 & 63.5 & 45.0 & 60.1 & \textbf{42.6} & \underline{65.7} & \textbf{45.2} & \textbf{17.9} & \textbf{49.6} \\ 
    IDOL\textsuperscript{\dag}~\cite{IDOL} & Pixel & 62.7 & 86.4 & 68.8 & \underline{54.9} & \underline{67.6} & \underline{59.1} & \textbf{82.9} & \underline{64.5} & \textbf{46.9} & \textbf{63.9} & 40.9 & \textbf{65.9} & \underline{42.9} & \underline{17.6} & 48.2 \\ 
    PM-VIS                              & Pixel&\underline{63.0}&\underline{86.9}&\underline{70.0}&54.5&66.9&\textbf{59.2}&\underline{82.3}&\textbf{64.9}&\underline{46.5}&\underline{63.0}&\underline{41.0}&65.6&41.7&\underline{17.6}&\underline{48.3}\\ \hline
    MaskFreeVIS~\cite{maskfreeVIS} & Box   & 55.3 & 82.5 & 60.8 & 50.7 & 62.2 & -    & -    & -    & -    & -    & -    & -    & -    & -   & -    \\
    IDOL-BoxInst                      & Box   & \underline{56.5} & \underline{83.3} & \underline{64.1} & \underline{49.7} & \underline{61.3} & \underline{53.2} & \underline{79.7} & \underline{59.6} & \underline{43.0} & \underline{58.1} & \underline{32.2} & \underline{55.7} & \underline{31.9} & \underline{15.8} & \underline{38.5}  \\
    \textbf{PM-VIS} & Box   & \textbf{59.7} & \textbf{84.8} & \textbf{67.7} & \textbf{52.3} & \textbf{64.3} &\textbf{55.8}&\textbf{80.6}&\textbf{61.8}&\textbf{44.4}&\textbf{60.2}&\textbf{37.5}&\textbf{62.6}&\textbf{37.5}&\textbf{16.8}&\textbf{43.9}\\ \hline
    \end{tabular}
\end{table*}

\subsubsection{Effect of the BoxInstLoss weight (W1)}
Although the pseudo masks we use provide a representation of the target within a specific range, they still exhibit limitations in accurately delineating the target. Issues such as significant pixel color variations across different parts of the target and color similarities among distinct targets pose challenges to the precision of target mask annotations. To address this problem, we introduce BoxInstLoss, a technique that leverages inter-pixel information. On one hand, this approach compensates for the incompleteness in capturing target boundaries. On the other hand, it addresses imprecision in target mask annotation by capitalizing on color similarities between pixels.
In Table~\ref{table:masklossW1}, we investigate the impact of the weight W1 of BoxInstLoss on VIS AP when training the model with pseudo masks. 
The best result is achieved at W1 = 1, while setting W1 = 0 means that the algorithm does not utilize BoxInstLoss. 
Our algorithm achieves a 2.1\% increase in AP, reaching 48.7\%, when using BoxInstLoss.

\subsubsection{Effect of the MaskLoss weight (W2)}
Table~\ref{table:masklossW2} demonstrates the influence of the weight (W2) assigned to MaskLoss on the algorithm. The optimal result is achieved with W2~=~0.5, whereas setting W2~=~0 means excluding MaskLoss and pseudo masks. The results indicate a noteworthy enhancement in performance with the incorporation of pseudo masks, resulting in a 3.9\% increase in AP, elevating it to 48.7\%. 
While BoxInstLoss, built upon bounding boxes, partly simulates real mask effects, this weakly supervised approach lacks object shape information and can be susceptible to object color variations. Consequently, its performance distinctly deviates from ground-truth conditions. 
Certainly, our proposed pseudo masks, by providing a rough contour of the object, enhance the algorithm's discernment of object characteristics such as shape, position, and size, thereby contributing to the algorithm's overall performance.

\subsubsection{Effect of pseudo masks for box-supervised VIS} 
Table~\ref{table:vspseudomasks} presents a comparison of various pseudo masks under the PM-VIS algorithm. 
It is evident from the AP performances of the three pseudo masks variants in the PM-VIS algorithm that Track-masks-final exhibits superior quality, achieving 48.7\% AP, while HQ-SAM-masks performs less favorably, attaining 46.8\% AP. 
For HQ-SAM-masks and IDOL-BoxInst-masks, we effectively address the issues of overlap and out-of-bounds in the initial pseudo masks using the DOOB strategy. When applied to IDOL-BoxInst-masks, this strategy improves the AP performance by 1.2\%, reaching 46.9\%.
When considering the highest-quality pseudo masks, Track-masks-final, the application of the SHQM method for single-keyframe tracking does not result in an improved algorithm performance. In fact, it leads to a 0.8\% decrease in AP. This outcome can be attributed to the predominance of IDOL-BoxInst-masks and HQ-SAM-masks in terms of quantity. The SHQM method, when selecting pseudo masks, appears to be misled by these lower-quality pseudo masks, inadvertently introducing a surplus of unnecessary errors.
Conversely, in the context of multi-keyframe tracking, the resultant pseudo masks consist of multiple sets of relatively high-quality masks. When selecting pseudo masks using the SHQM method in this scenario, it introduces a significant number of better-quality tracking masks. This leads to a 0.5\% relative increase in AP compared to single-frame scenarios, ultimately reaching an optimal level of 48.7\% AP.

\subsubsection{Effect of pseudo masks for fully supervised VIS} 
Among the two types of pseudo masks with an equal number of instance annotations, IDOL-BoxInst-masks and Track-masks-final, the annotation quality of IDOL-BoxInst-masks is observed to be lower than that of Track-masks-final, as evident from the results of weak supervision. In some cases, there are instances that are expected to be recognized or segmented more effectively by weak supervision methods, but the results in IDOL-BoxInst-masks are less satisfactory. 
On the other hand, Track-masks-final recognizes fewer of these instances. As shown in Table~\ref{table:twopseudomasksmaskIOUgTvs}, we conduct a comparative analysis of the results obtained when using these two types of pseudo masks as aids in filtering ground-truth data. The results show that training with PM-VIS on ground-truth data filtered with the assistance of Track-masks-final outperforms IDOL-BoxInst-masks by 1.3\%, achieving an AP of 50.0\%.

\subsubsection{Effect of BoxInstLoss for fully supervised VIS} 
Table~\ref{table:BoxInstLossforfullysup} presents the performance of our PM-VIS algorithm compared to the IDOL algorithm without BoxInstLoss on ground-truth data. The training set used here comprises YTVIS2019 ground-truth data that has undergone two filtering methods: Missing-Data and RIA. These results showcase the algorithm performance on YTVIS2019 VAL. The introduction of BoxInstLoss in our algorithm is primarily motivated by its ability to examine instance segmentation from a pixel similarity perspective. Therefore, we believe that it can still be effective in a fully supervised scenario. As observed, with the inclusion of BoxInstLoss, our algorithm demonstrates a notable 2.3\% increase in AP, reaching 50.0\%.

\subsubsection{Effect of Missing-Data and RIA} 
Table~\ref{table:missingdatamaskIOUgT} illustrates the impact of two filtering and optimization methods, Missing-Data and RIA, on algorithm performance. It's evident that the current algorithms struggle to handle the issues present in the filtered-out data. 
When training the IDOL-BoxInst model using all ground-truth data without any pixel-level annotations, we notice that the algorithm AP decreases by 0.9\% compared to when the missing data is not used. 
Training the PM-VIS model on ground-truth data filtered only by Missing-Data leads to a slight improvement in AP, with a 0.2\% increase compared to using all ground-truth data. 
Furthermore, when training the PM-VIS model on ground-truth data filtered using both methods, the trained model achieves a significant increase in AP, improving by 0.9\% to reach 50.0\%.


\subsubsection{Effect of various mask IoU thresholds ($\tau$) on the RIA} 
Table~\ref{table:maskIOUgT} presents the impact of different mask IoU thresholds on RIA regarding algorithm performance. These thresholds are determined by calculating mask IoU between Track-masks-final and ground-truth data. When $\tau$ is set to 0, it indicates that this filtering method is not applied, whereas $\tau$ greater than 0 signifies the removal of instance annotations from ground-truth data when their mask IoU values fall below $\tau$. It is worth to note that the mask IoU is computed between these pseudo masks and ground-truth data because lower mask IoU values correspond to lower mask quality. Despite employing various fusion strategies, enhancing the quality of these instance pseudo masks proves challenging. Consequently, we can conclude that these instances are difficult to identify in certain video frames. Removing them has the potential to improve algorithm performance. In our experiments, setting $\tau$ to 0.6 leads to a 0.7\% improvement in the algorithm AP compared to not using this filtering method, achieving a performance level of 50.0\%.

\begin{figure*}[!ht]
	\centering
	\scalebox{0.44}{\includegraphics{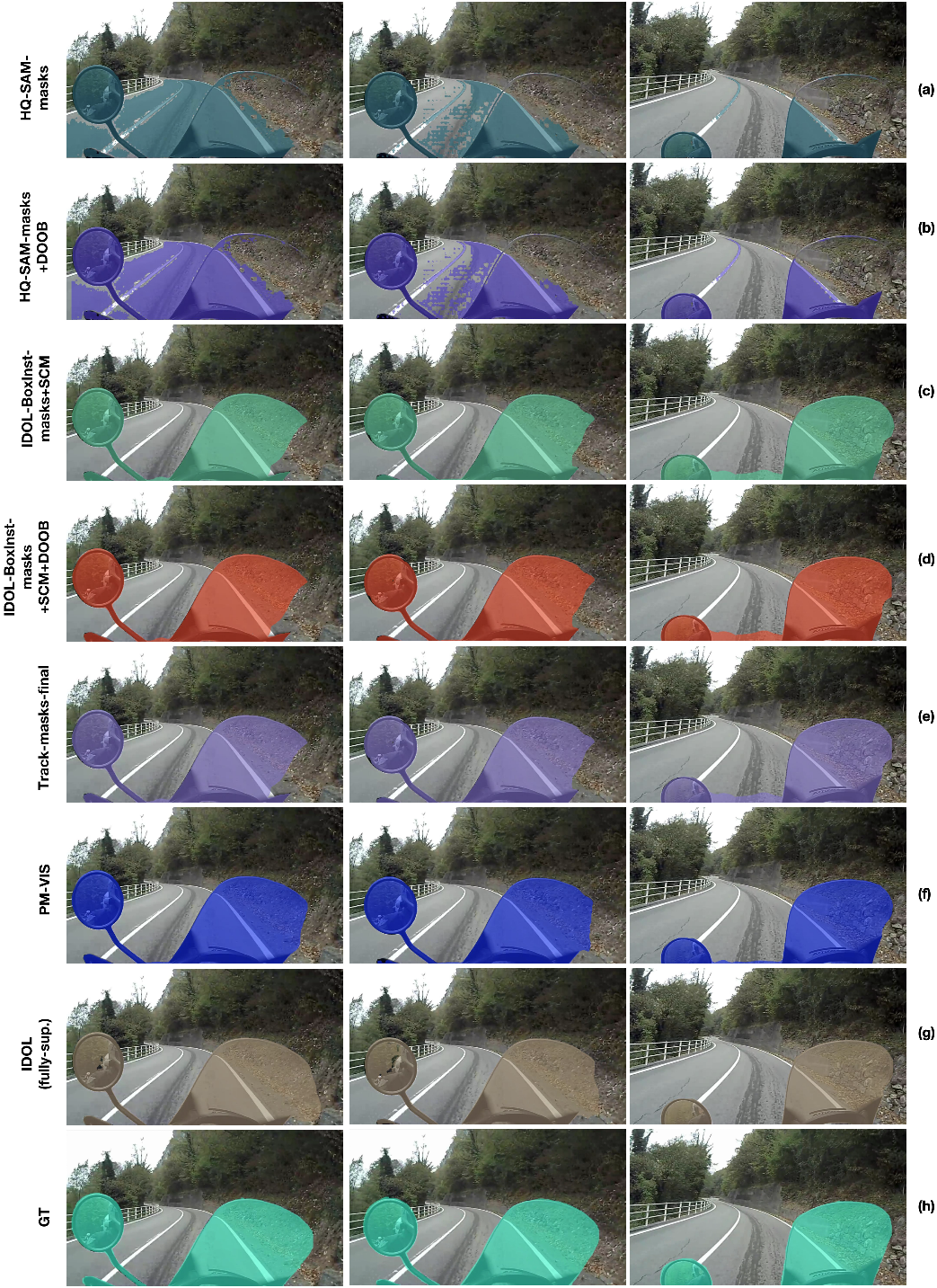}}
	\caption{Visualization of instance masks in the YTVIS2019 training set. Rows (a) and (b) depict the predicted pseudo masks generated using the HQ-SAM model with gtbox as proposals. Rows (c) and (d) display the pseudo masks predicted by the IDOL-BoxInst model. Row (e) represents Track-masks-final used for training the weakly supervised PM-VIS model. Row (f) displays the results predicted by the weakly supervised PM-VIS model. Rows (g) and (h) show the results predicted by the fully supervised VIS model (IDOL) and the gtmasks, respectively.}
	\label{fig:pseudomasksVs} 
\end{figure*}

\subsection{Comparison with SOTA methods}
We have compared the PM-VIS model trained without using the gtmasks with SOTA weakly supervised methods using ResNet-50 and Swin Large backbones on YTVIS2019, YTVIS2021, and OVIS datasets. Furthermore, we compare the performance of the PM-VIS model trained using the filtered ground-truth data with other fully supervised VIS algorithms.
The results are presented in Table~\ref{table:sotaResNet50} and Table~\ref{table:sotaswinL}.

\subsubsection{ResNet-50 backbone} 
As shown in Table~\ref{table:sotaResNet50}, it is apparent that with the ResNet-50 backbone, our basic IDOL-BoxInst model, trained solely on box annotations, underperforms MaskFreeVIS~\cite{maskfreeVIS} by a margin of 2.7\% AP on YTVIS2019. 
Upon the incorporation of pseudo masks into the PM-VIS algorithm, we observe a noteworthy 4.8\% improvement in AP compared to IDOL-BoxInst. Moreover, this enhancement outperforms the performance of the only box-supervised VIS algorithm, MaskFreeVIS, by 2.1\%, achieving an outstanding AP of 48.7\%.
Similarly, on the YTVIS2021 and OVIS datasets, our IDOL-BoxInst algorithm achieves competitive results, surpassing the MaskFreeVIS and even outperforming some fully supervised VIS algorithms.
After integrating pseudo masks into the PM-VIS algorithm, we surpass all box-supervised VIS algorithms, outperforming MaskFreeVIS by 3.7\% and 12.1\%, thus establishing a new SOTA. 

Furthermore, training the PM-VIS model using ground-truth data filtered with pseudo masks also yields improved results for three primary reasons. Firstly, the BoxInstLoss enhances mask segmentation precision by extending constraints to both bounding box ranges and pixel similarity. Secondly, a thorough examination, including visual inspection and model performance analysis, demonstrates that the quality of the acquired pseudo masks (Track-masks-final) is approaching that of gtmasks. Thus, the main factor affecting weakly supervised performance appears to be the presence of low-quality instance mask annotations. The application of our proposed filtering methods (Missing-Data and RIA) retains higher-quality, less ambiguous data. Finally, especially for models employing backbone with limited feature extraction capabilities like ResNet-50, the presence of distinct instance features allows the algorithm to efficiently and accurately acquire necessary information, minimizing the need for extensive feature learning and resulting in faster, more precise utilization of the acquired information.
When training the PM-VIS model on the filtered ground-truth data, the AP surpasses IDOL\textsuperscript{\dag} on all three datasets, achieving 50.0\%, 47.7\%, and 29.9\% on each, respectively. It's evident that our collection of pseudo masks not only enhances the performance of box-supervised VIS but also contributes to an improvement in fully supervised VIS to a certain extent.

\subsubsection{Swin Large backbone}
As presented in Table~\ref{table:sotaswinL}, we also conduct experiments using Swin-L as the backbone on YTVIS2019, YTVIS2021 and OVIS. 
The comparison reveals that our algorithm PM-VIS outperforms the basic IDOL-BoxInst by 3.2\%, 2.6\%, and 5.3\% on the respective datasets. Similarly, we also surpass MaskFreeVIS by 4.4\% on the YTVIS2019 dataset, establishing a new SOTA for box-supervised VIS algorithms.
The outstanding performance significantly narrows the performance gap between fully supervised and weakly supervised VIS methods.

Furthermore, in theory, training the PM-VIS model with filtered ground-truth data will also result in performance exceeding IDOL\textsuperscript{\dag} with the Swin-L backbone. 
As our analysis during the training of models with ResNet-50 suggests, stronger backbones, such as Swin-L, possess a greater ability to extract information from ambiguous or low-quality instance segmentation annotations. 
When utilizing the Swin-L backbone, the model will extract target information from the data more precisely. This implies that by using less data, specifically excluding non-essential data from the dataset, the model can still achieve or even surpass the performance obtained by using the entire dataset.
We train the PM-VIS model using Swin-L backbone on the same filtered ground-truth data used for training with ResNet-50 backbone, and the model consistently delivers the expected performance.
Specifically, on the YTVIS2019, YTVIS2021, and OVIS datasets, the fully supervised VIS model, PM-VIS, outperforms IDOL\textsuperscript{\dag} in terms of AP, achieving 63.0\%, 59.2\%, and 41.0\%, respectively. This illustrates that our pseudo masks enhance the performance of weakly supervised algorithms and lead to improvements in fully supervised scenarios, regardless of whether a stronger or weaker backbone is employed.

\subsection{Visualization}
Fig.~\ref{fig:pseudomasksVs} illustrates different mask visualization examples. HQ-SAM-masks generated by HQ-SAM shows poor segmentation quality and noticeable errors, making them less effective compared to other pseudo masks. IDOL-BoxInst-masks, assisted by HQ-SAM-masks, demonstrates improved quality and accuracy by aligning with corresponding gtboxes and better capturing target shapes, as shown in the figure. Track-masks-final represents a high-quality pseudo mask collection selected from three sources: HQ-SAM-masks, IDOL-BoxInst-masks, and the original Track-masks.
PM-VIS employs pseudo mask information from Track-masks-final, refining it through training to effectively enhance recognition accuracy. It's worth noting that from the figures, we can observe that the predictions of our weakly supervised algorithm PM-VIS closely resemble those of the gtmasks.


\section{Conclusion}
Despite the substantial performance improvement, contemporary VIS algorithms encounter the constraint of high costs in annotating for instances and inadequate training data for real-world applications. As a result, weakly supervised methods have emerged. 
While the single-step methods, which solely utilize box annotations for model training, have achieved notable results, there still exists a significant gap compared to fully supervised methods.
In this paper, we introduce a novel two-step approach that leverages high-performance models to generate instance pseudo masks and trains the VIS model with these pseudo masks and box annotations.
Firstly, we introduce three effective methods for generating target pixel-level mask information utilizing HQ-SAM, IDOL-BoxInst and DeAOT models based on object box annotations. Secondly, we introduce three pseudo mask refinement methods, namely SCM, SHQM, and DOOB, designed to achieve higher-quality pseudo masks based on model characteristics and pseudo mask attributes. Subsequently, utilizing the training dataset improved with these pseudo masks, we train the proposed PM-VIS model without using gtmasks, which demonstrates SOTA performance in YTVIS2019, YTVIS2021 and OVIS. Furthermore, by capitalizing on the characteristics of pseudo masks, we introduce the use of pseudo masks as auxiliary tools for filtering ground-truth data. These filtering methods, named Missing-Data and RIA, are employed to eliminate challenging instance annotations. As a result, when the PM-VIS model is trained on the filtered ground-truth dataset, it exhibits a significant improvement in recognition performance over the baseline. We believe that our contributions and insights will serve as valuable inspiration for future applications and research in both fully supervised and weakly supervised approaches.

\bibliographystyle{IEEEtran}  
\bibliography{egbib.bib}

\end{document}